%% file: dopelearning.tex
\documentclass{sig-alternate-05-2015}

\usepackage[ruled, vlined, nofillcomment]{algorithm2e}

\usepackage{url}
\usepackage{amsmath}
\usepackage{amssymb}
\usepackage{graphicx}
\usepackage[tight]{subfigure}
\usepackage[english]{babel}
\usepackage{booktabs}
\usepackage{enumitem}
\usepackage{xspace}
\usepackage{tipa}
\usepackage{verse}
\usepackage{alltt}
\usepackage{color, array}
\usepackage{hyphenat}

\usepackage{tikz}
\usetikzlibrary{shapes,positioning,fit,calc}

\setlist[description]{leftmargin = 0pt, parsep = 0pt}

\input{defines}

\toappear{This is a pre-print of an article appearing at KDD'16.}

\begin{document}


\title{DopeLearning: A Computational Approach to Rap Lyrics 
Generation\thanks{When used as an adjective, 
{\em dope} means {\em cool}, {\em nice}, or {\em awesome}.}}

\numberofauthors{5}

\author{
\alignauthor Eric Malmi \\
\affaddr Aalto University and HIIT\\
\affaddr Espoo, Finland \\
\email{eric.malmi@aalto.fi}
\alignauthor Pyry Takala \\
\affaddr Aalto University \\
\affaddr Espoo, Finland \\
\email{pyry.takala@aalto.fi}
\alignauthor Hannu Toivonen \\ 
\affaddr University of Helsinki and HIIT\\
\affaddr Helsinki, Finland \\
\email{hannu.toivonen@cs.helsinki.fi}
\and
\alignauthor Tapani Raiko \\
\affaddr Aalto University \\
\affaddr Espoo, Finland \\
\email{tapani.raiko@aalto.fi}
\alignauthor Aristides Gionis \\
\affaddr Aalto University and HIIT \\
\affaddr Espoo, Finland \\
\email{aristides.gionis@aalto.fi}
}

\date{\today}


\maketitle
\begin{abstract}
Writing rap lyrics requires both creativity to construct a meaningful, 
interesting story and lyrical skills to produce complex rhyme patterns,  
which form the cornerstone of good flow. 
We present a rap lyrics generation method that captures both of these aspects. 
First, we develop a prediction model to identify the next line of existing 
lyrics from a set of candidate next lines. This model is based on two 
machine-learning techniques: 
the RankSVM algorithm and a deep neural network model with a novel structure. 
Results show that the prediction model can identify the true next line 
among 299 randomly selected lines with an accuracy of 17\%, i.e., over 50 times 
more likely than by random.
Second, we employ the prediction model to combine lines from existing songs, 
producing lyrics with rhyme and a meaning.
An evaluation of the produced lyrics shows that in terms of quantitative rhyme 
density, the method outperforms the best human rappers by 21\%.
The rap lyrics generator has been deployed as an online tool called \deepbeat, 
and the performance of the tool has been assessed by analyzing its
usage logs. This analysis shows that machine-learned rankings correlate 
with user preferences.
\end{abstract}

\section{Introduction}


Emerging from a hobby of African American youth in the 1970s, rap music has 
quickly 
evolved into a mainstream music genre with several artists frequenting
Billboard top rankings.
Our objective is to study the problem of computational creation of rap lyrics.
Our interest in this problem is motivated by two different different 
perspectives. 
First, we are interested in analyzing the formal structure of rap lyrics
and in developing a model that can lead to generating artistic work.
Second, with the number of smart devices increasing that we use on a daily 
basis, it is expected that the demand will increase for systems that interact 
with humans in non-mechanical and pleasant ways.

Rap is distinguished from other music genres by the formal structure present in 
rap lyrics, which makes the lyrics rhyme well and hence provides better flow 
to the music. Literature professor Adam Bradley 
compares rap with popular music and traditional poetry, stating that while 
popular lyrics lack much of the formal structure of literary verse, rap crafts 
``intricate structures of sound and rhyme, creating some of the most 
scrupulously formal poetry composed today'' \cite{bradley2009}.

We approach the problem of lyrics creation from an 
information\hyp{}retrieval (IR) perspective. 
We assume that we have access to a large repository of rap-song lyrics. 
In this paper, we use a dataset 
containing over half a million lines from lyrics of
104 different rap artists.
We then view the lyrics\hyp{}generation problem
as the task of identifying a {\em relevant next line}.
We consider that a rap song has been partially constructed
and treat the first $m$ lines of the song as a \emph{query}.
The IR task is to identify the most relevant next line from a collection of 
candidate lines, with respect to the query. 
Following this approach, new lyrics are constructed line by line, 
combining lyrics from different artists in order to introduce novelty. 
A key advantage of this approach is that we can evaluate the performance of the generator by 
measuring how well it predicts existing songs. 
While conceptually one could approach the lyrics-generation problem
by a word-by-word construction, 
so as to increase novelty, 
such an approach would require 
significantly more complex models
and we leave it for future work.

Our work lies in the intersection between the areas of 
{\em computational creativity}
and {\em information retrieval}.
In our approach,
we assume that users have a certain concept in their mind, 
formulated as a sequence of rap lines, 
and their information need is to find the missing lines, composing a song.
Such an information need does not have a factual answer; 
nevertheless, users will be able to assess the relevance of the response 
provided by the system. 
The relevance of the response depends on factors that include
rhyming, 
vocabulary, 
unexpectedness, 
semantic coherence, and
humor.
Tony Veale \cite{veale2011creative} illustrates other linguistically creative 
uses of information retrieval, e.g., for metaphor generation. 
He argues that phrases extracted from large corpora can be used 
as ``readymade'' or ``found'' objects, like \emph{objets trouv\'es} in arts, 
that can take on fresh meanings when used in a new context.

From the computational perspective, a major challenge in generating rap lyrics 
is to produce semantically coherent lines instead of merely
generating complex rhymes. As Paul Edwards~\cite{edwards2009} puts it: ``If 
an 
artist takes his or her time to craft phrases that rhyme in intricate ways but 
still gets across the message of the song, that is usually seen as the mark of a 
highly skilled MC.''\footnote{MC, short for master of ceremonies or microphone 
controller, is essentially a word for a rap artist.} 
As a result of record results in computer vision, deep neural networks 
\cite{Bengio-et-al-2015-Book} have become a popular tool for feature learning.
To avoid hand-crafting a number of semantic and grammatical features, we 
introduce a deep neural network model that maps sentences into a 
high-dimensional vector space. This type of  vector-space representations have 
attracted much attention in recent years and have exhibited great empirical 
performance in tasks requiring semantic analysis of natural 
language~\cite{mikolov2013,pennington2014}.

While some of the features we extract from the analyzed lyrics are tailored for 
rap lyrics, a similar approach could be applied to generate lyrics for other 
music genres.
Furthermore, the proposed framework could form the basis for 
several other text-synthesis problems, such as generation of text or conversation
responses. Practical extended applications include automation of tasks,
such as customer service, sales, or even news reporting. 

Our contributions can be summarized as follows:

\squishlist
\item[($i$)] We propose an information-retrieval approach to rap lyrics generation. A 
similar approach could be applied to other tasks requiring text synthesis.

\item[($ii$)] We introduce several useful features for predicting the next line of a 
rap song and hence for generating new lyrics. In particular, we have developed 
a deep neural network model for capturing the semantic similarity of lines. 
This feature carries the most predictive power of all the features we have 
studied.

\item[($iii$)] We present \emph{rhyme density}, a measure for the technical quality of 
rap lyrics. This measure is validated with a human subject, a native-speaking 
rap artist.

\item[($iv$)] We have built an online demo of the rap lyrics generator openly 
available at \website. The performance of the algorithm has been assessed based 
on the usage logs of the demo.

\squishend


The rest of the paper is organized as follows. We start with a brief discussion of 
relevant work. Next, we discuss the domain of rap lyrics, introduce the dataset
that we have used, and describe how rhymes can be analyzed. We proceed by 
describing the task of \nextline and our approach to solving it, including the 
used features, the neural language model, and the results of our experiments. 
We apply the resulting model to the task of lyrics generation, showing also 
examples of generated lyrics. The final sections discuss further these tasks, 
our model, and conclusions of the work.

\section{Related work}

While the study of human-generated lyrics is of interest to academics in
fields such as linguistics and music, artificial rhyme generation is 
also relevant for various subfields of computer science. Relevant literature 
can be found
under domains of computational creativity, information extraction and
natural language processing. Additionally, relevant methods can be found 
under the domain of machine learning, for instance, in the emerging field of 
deep learning.

Hirjee and Brown \cite{hirjee2009,hirjee2010} develop a probabilistic method, 
inspired by 
local alignment protein homology detection algorithms, for detecting rap 
rhymes. Their algorithm obtains a high rhyme detection performance, but it 
requires a training dataset with labeled rhyme pairs. We introduce a simpler, 
rule-based approach in Section~\ref{sec:raplyzer}, which seemed sufficient 
for our purposes. Hirjee and Brown obtain a phonetic transcription for the 
lyrics by applying the CMU Pronouncing Dictionary~\cite{cmudict}, some 
hand-crafted rules to handle slang words, and text-to-phoneme rules to handle 
out-of-vocabulary words, whereas we use an open-source speech synthesizer, 
eSpeak, to produce the transcription. The computational generation of rap 
lyrics has been previously studied in \cite{wu2013,wu2015}. These works adopt a 
machine-translation approach, whereas we view it as an information-retrieval 
problem. Furthermore, we have deployed our lyrics generator as an 
openly accessible web tool to assess its performance in the wild. 

Automated creation of rap lyrics can also be viewed as a problem within
the research field of Computational Creativity,
i.e., study of computational systems
which exhibit behaviors deemed to be creative~\cite{colton2012computational}.
According to Boden~\cite{Boden2004},
creativity is the ability to come up with ideas or artifacts
that are new, surprising, and valuable, and there are 
three different types of creativity.
Our work falls into the class of
``combinatorial creativity'' where creative results are
produced as novel combinations of familiar ideas.
Combinatorial approaches have been used to create poetry before
but were predominantly based on the idea of copying the grammar from existing poetry
and then substituting content words with other ones~\cite{toivanenetal2012}.

In the context of web search, it has been shown that document ranking 
accuracy can be significantly improved by combining multiple features 
via machine-learning algorithms instead of using a single static ranking, 
such as Page\-Rank~\cite{richardson2006}. This \emph{learning-to-rank} approach is very 
popular nowadays and many methods have been developed for it~\cite{liu2009}. 
In this paper, we use the RankSVM algorithm \cite{joachims2002} for 
combining different relevance features for the next-line prediction problem, 
which is the basis of our rap-lyrics generator.


Neural networks have been recently applied to various related tasks. For instance,
recurrent neural networks (RNNs) have shown promise in predicting text sequences
\cite{Graves13,ICML2011Sutskever_524}.
Other applications include tasks such as information extraction, information retrieval and indexing
\cite{Deng11}. Question answering has also been approached by using deep
learning to map questions and answers to a latent semantic space, and then 
either generating a response \cite{Oriol} or selecting one \cite{Yu:2014}. Our 
neural network approach has some similarity to response selection as we also 
learn a mapping to a hidden space. On the other hand, our network architecture 
uses a feed-forward net---a suitable choice in our context where sentences are 
often relatively short and equal in length. Also, generation of lyrics is not 
typically considered as a response-selection task but we interpret it as one.

\section{Anatomy of rap lyrics}

In this section, we first describe a typical structure of rap lyrics, 
as well as different rhyme types that are often used. 
This information will be the basis for extracting useful features for the next-line prediction problem, 
which we discuss in the next section. 
Then we introduce a method for automatically detecting rhymes, 
and we define a measure for rhyme density.
We also present experimental results to assess the validity of the rhyme-density measure.

\subsection{Rhyming}

Various different rhyme types, such as \emph{perfect rhyme}, \emph{alliteration}, and 
\emph{consonance}, are employed in rap lyrics, but the most common rhyme type nowadays, 
due to its versatility, is the \emph{assonance} rhyme \cite{rapgenius,edwards2009}. 
In a perfect rhyme, the words share exactly the same end sound, as in ``slang 
-- gang,'' whereas in 
an assonance rhyme only the vowel sounds are shared.
For example, words 
``crazy'' and ``baby'' have different consonant sounds, but the vowel sounds 
are the same as can be seen from their phonetic representations 
``\textipa{k\*r\underline{eI}s\underline{i}}'' and 
``\textipa{b\underline{eI}b\underline{i}}.''

An assonance rhyme does not have to cover only the end of the rhyming words but 
it can span multiple words as in the example below (rhyming part is 
highlighted). This type of assonance rhyme is called \emph{multisyllabic rhyme}.
\begin{center}
\begin{tabular}{l}
\small{``This is a job --- I get paid to {\bf sling some raps},} \\
\small{What you made last year was less than my {\bf income tax}''} 
\cite{flocabulary}
\end{tabular}
\end{center}
It is stated in \cite{flocabulary} that ``[Multisyllabic rhymes] are 
hallmarks of all the dopest flows, and all the best rappers use them,''


\subsection{Song structure}

A typical rap song follows a pattern of alternating verses and choruses. These
in turn consist of lines, which break down to individual words and finally to 
syllables. A line equals one musical bar, which typically consists of four 
beats, setting limits to how many syllables can be fit into a single line. 
Verses, which constitute the main body of a song, are often composed of 16 
lines. \cite{edwards2009}

Consecutive lines can be joined through rhyme, which is typically placed at the 
end of the lines but can appear anywhere within the lines. The same end rhyme 
can be maintained for a couple of lines or even throughout an entire 
verse. In the verses our algorithm generates, the end rhyme is kept fixed for 
four consecutive lines unless otherwise specified by the user (see 
Appendix~\ref{sec:verses} for an example).

\subsection{Automatic rhyme detection} \label{sec:raplyzer}

Our aim is to automatically detect multisyllabic assonance rhymes from lyrics 
given as text.
For this, we first obtain a phonetic transcription of the lyrics by applying the 
text-to-phonemes functionality of an open source speech synthesizer
eSpeak.\footnote{\url{http://espeak.sourceforge.net/}} The synthesizer
assumes a typical American--English pronunciation. From the phonetic 
transcription, we can detect rhymes by finding matching vowel phoneme sequences, 
ignoring consonant phonemes and spaces.

\subsubsection{Rhyme density measure} \label{sec:rhymedensity}

In order to quantify the technical quality of lyrics from a rhyming 
perspective, we introduce a measure for the \emph{rhyme density} of the lyrics. 
A simplified description\footnote{For the details, see: 
\url{https://github.com/ekQ/raplysaattori}} for the computation of this measure 
is provided below:
\squishlist
 \item[1.] Compute the phonetic transcription of the lyrics and remove all but 
vowel phonemes.
 \item[2.] Scan the lyrics word by word.
 \item[3.] For each word, find the longest matching vowel sequence 
(=multisyllabic assonance rhyme) in the proximity of the word.
 \item[4.] Compute the rhyme density by averaging the lengths of the longest 
matching vowel sequences of all words.
\squishend

The rhyme density of an artist is computed as the average rhyme density of his 
or her (or its) songs.
Intuitively speaking, rhyme density means the average length of the longest 
rhyme per word.


\subsubsection{Data} \label{sec:data}

We compiled a list of 104 popular English-speaking rap artists and scraped 
all their songs available on a popular lyrics website. In total, we have 
583\,669 lines from 10\,980 songs.

To make the rhyme densities of different artists comparable, we normalize the 
lyrics by removing all duplicate lines within a single song, 
as in some cases the lyrics contain the chorus repeated many times, whereas in 
other cases they might just have ``Chorus 4X,'' depending on the user who has 
provided the lyrics. Some songs, like intro tracks, often contain more regular 
speech rather than rapping, and hence, we have removed all songs whose title 
has one of the following words: ``intro,'' ``outro,'' ``skit,'' or 
``interlude.''

\subsubsection{Evaluating human rappers' rhyming skills}

We initially computed the rhyme density for 94 rappers, ranked the artists 
based on this measure, and published the results online \cite{raplyzer}.
An excerpt of the results is shown in 
Table~\ref{tab:densities}.

\begin{table}[t]
\caption{A selection of popular rappers and their rhyme 
densities, i.e., their average rhyme lengths per word.}\label{tab:densities}
\centering
\begin{tabular}{lll}
\toprule
Rank & Artist & Rhyme density \\
\midrule
1. & Inspectah Deck & 1.187 \\
2. & Rakim   & 1.180 \\
3. & Redrama & 1.168 \\
30. & The Notorious B.I.G.    & 1.059 \\
31. & Lil Wayne   & 1.056 \\
32. & Nicki Minaj & 1.056 \\
33. & 2Pac    & 1.054 \\
39. & Eminem  & 1.047 \\
40. & Nas & 1.043 \\
50. & Jay-Z   & 1.026 \\
63. & Wu-Tang Clan    & 1.002 \\
77. & Snoop Dogg  & 0.967 \\
78. & Dr. Dre & 0.966 \\
94. & The Lonely Island   & 0.870 \\
\bottomrule
\end{tabular}
\end{table}

Some of the results are not too surprising; for instance Rakim, who is ranked second, 
is  known for ``his pioneering use of internal rhymes and multisyllabic 
rhymes.''\footnote{The {W}ikipedia article on {R}akim 
\url{http://en.wikipedia.org/wiki/Rakim} (Accessed: 2016-02-11)} 
On the other hand, a limitation of the results is that 
some artists, like Eminem, who use a lot of multisyllabic rhymes but construct 
them often by bending words (pronouncing words unusually to make them rhyme), 
are not as high on the list as one might expect.

\begin{figure}[t]
  \centering
  \includegraphics[width=0.7\columnwidth]{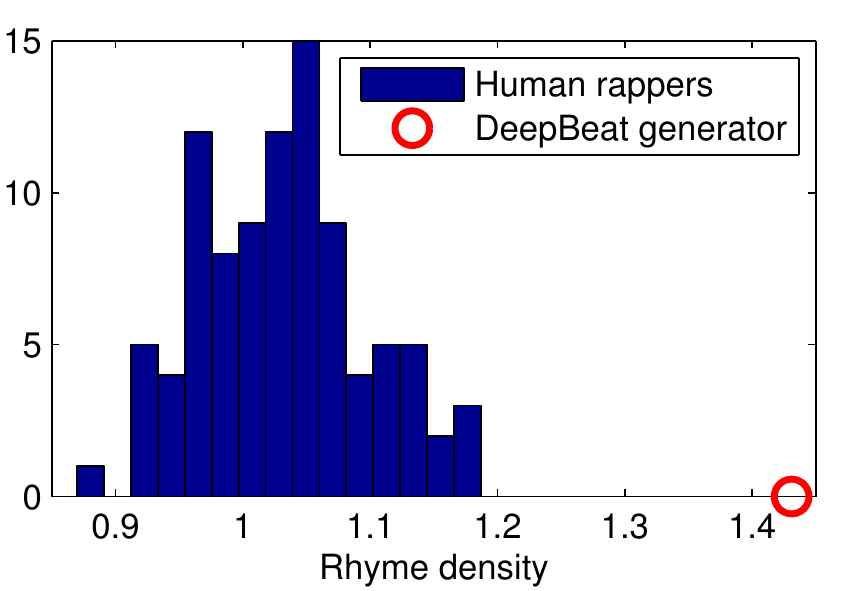}
  \caption{Rhyme density distribution of 105 rappers.}
  \label{fig:hist}
\end{figure}

In Figure~\ref{fig:hist}, we show the distribution of rhyme densities along 
with the rhyme density obtained by our lyrics generator algorithm \deepbeat 
(see Section~\ref{sec:res} for details).

\begin{table}[t]
\caption{List of rap songs ranked by the algorithm and by the artist 
himself according to how technical he perceives them. Correlation is
0.42.}\label{tab:ahmen}
\centering
\begin{tabular}{ccc}
\toprule
Rank by artist & Rank by algorithm & Rhyme density \\
\midrule
1. & 1. & 1.542 \\
2. & 4. & 1.214 \\
3.--4. & 9. & 0.930 \\
3.--4. & 3. & 1.492 \\
5. & 2. & 1.501 \\
6.--7. & 10. & 0.909 \\
6.--7. & 7. & 1.047 \\
8.--9. & 6. & 1.149 \\
8.--9. & 5. & 1.185 \\
10. & 8. & 1.009 \\
11. & 11. & 0.904 \\
\bottomrule
\end{tabular}
\end{table}

\subsection{Validating the rhyme density measure}

After we had published the results shown in Table~\ref{tab:densities} online, 
a rap artist called \emph{Ahmen} contacted us, asking us to compute the rhyme density for the
lyrics of his debut album. Before revealing the rhyme densities of the 11 
individual songs he sent us, we asked the rapper to rank his own lyrics 
``starting from the most technical according to where you think you have used 
the most and the longest rhymes.'' The rankings produced by the artist and by 
the algorithm are shown in Table~\ref{tab:ahmen}.

We can compute the correlation between the artist produced and algorithm 
produced rankings by applying the Kendall tau rank correlation coefficient. 
Assuming that all ties indicated by the artist are decided unfavorably for the 
algorithm, the correlation between the two rankings would still be 0.42, and 
the null hypothesis of the rankings being independent can be 
rejected ($p<0.05$). This suggests that the rhyme density measure adequately 
captures the technical quality of rap lyrics.

\section{Next line prediction} 
\label{sec:nextline}

We approach the problem of rap lyrics generation as an information-retrieval task.
In short, our approach is as follows: 
we consider a large repository of rap lyrics, 
which we treat as training data, 
and which we use to learn a model between consecutive lines in rap lyrics. 
Then, given a set of seed lines
we can use our model to identify the best next line 
among a set of candidate next lines taken from the lyrics repository.
The method can then be used to construct a song line-by-line, 
appending relevant lines from different songs. 

In order to evaluate this information-retrieval task, 
we define the problem of \emph{``next-line prediction.''}

\subsection{Problem definition}

As mentioned above, 
in the core of our algorithm for generating rap lyrics 
is the task of finding the next line of a given song.
This next line prediction problem
is defined as follows.

\begin{problem}{\em (}\nextline{\em )}
Consider the lyrics of a rap song $S$, 
which is a sequence of $n$ lines $(s_1, \ldots, s_n)$.
Assume that the first $m$ lines of $S$, denoted by $B=(s_1, \ldots, s_m)$, 
are known and are considered as ``the query.''
Consider also that we are given a set of $k$ {\em candidate next lines}
$C = \{\ell_1, \ldots, \ell_k\}$, 
and the {\em promise} that $s_{m+1}\in C$.
The goal is to identify $s_{m+1}$ in the candidate set $C$, 
i.e., pick a line $\ell_i\in C$ such that $\ell_i = s_{m+1}$.
\end{problem}

Our approach to solving the \nextline problem 
is to compute a relevance score between the query $B$ and each candidate line $\ell\in C$, 
and return the line that maximizes this relevance score. 
The performance of the method can be evaluated using standard information 
retrieval measures, such as mean reciprocal rank. 
As the relevance score we use a linear model over a set of similarity features 
between the query song-prefix $B$ and the candidate next lines $\ell\in C$. 
The weights of the linear model are learned using the \ranksvm\ algorithm, 
described in Section~\ref{sec:ranksvm}.

In the next section, we describe a set of features that we use 
for measuring the similarity between the 
previous lines $B$ and a candidate next line $\ell$.

\subsection{Feature extraction} 
\label{sec:feats}

The similarity features we use for the next-line prediction problem 
can be divided into three groups, capturing 
($i$) \emph{rhyming}, 
($ii$) \emph{structural similarity}, 
and 
($iii$) \emph{semantic similarity}.

($i$) We extract three different rhyme features based on the phonetic transcription 
of the lines, as discussed in Section~\ref{sec:raplyzer}.

\spara{\erhyme} is the number of matching vowel phonemes at the end of lines 
$\ell$ and $s_m$, i.e., the last line of $B$.
Spaces and consonant phonemes are ignored, so for instance, the 
following two phrases 
would have three end rhyme vowels in common.

\begin{center}
\begin{tabular}{ll}
\emph{Line} & \emph{Phonetic transcription} \\
pay for & \textipa{p\underline{e}\underline{I} f\underline{O}:\*r} \\
stay warm & \textipa{st\underline{e}\underline{I} w\underline{O}:\*rm} \\
\end{tabular}
\end{center}

\spara{\eerhyme} is the number of matching vowel phonemes at the end of lines 
$\ell$ and $s_{m-1}$, i.e., the line before the last in~$B$.
This feature captures alternating rhyme schemes of the form \emph{``abab.''}

\spara{\intrhyme} is the average number of matching vowel phonemes per word. 
For each word in $\ell$, we find the longest matching vowel sequence in 
$s_{m}$ and average the lengths of these sequences.
This captures other than end rhymes.

\smallskip
($ii$) With respect to the structural similarity between lines, 
we extract one feature measuring the difference of the lengths of the lines.

\spara{\linelength}.\ Typically, consecutive lines are roughly the same length 
since they need to be spoken out within one musical bar of a fixed length. 
The length similarity of two lines $\ell$ and $s$ is computed as
\begin{equation*}
 1- \frac{\abs{\len{\ell} - \len{s}}}
 {\max\left(\len{\ell}, \len{s}\right)},
\end{equation*}
where $\len{\cdot}$ is the function that returns the number of characters in a 
line.\footnote{We also tested the number of syllables in a line but the results 
were similar.}
We compute the length similarity between a candidate line $\ell$ and the last line $s_m$ 
of the song prefix $B$.


\smallskip
($iii$) Finally, for measuring semantic similarity between lines,
we employ four different features.

\spara{\bow}.\ 
First, we tokenize the lines and represent each last line $s_m$ of $B$ as a bag of words $S_m$.
We apply the same procedure and obtain a bag of words $L$ for a candidate line $\ell$.
We then measure semantic similarity between two lines
by computing the Jaccard similarity between the corresponding bags of words
\begin{equation*}
\frac{\abs{S_m \cap L}}{\abs{S_m \cup L}}.
\end{equation*}

\spara{\bowfive}.\ 
Instead of extracting a bag-of-words representation from only the last line $s_m$ of $B$, 
we use the $k$ previous lines.
\begin{equation*}
\frac{\abs{\left(\bigcup_{j=m-k}^m S_j\right) \cap L}}
{\abs{\left(\bigcup_{j=m-k}^m S_j\right) \cup L}}.
\end{equation*}
In this way we can incorporate a longer context that could be relevant
with the next line that we want to identify.
We have experimented with various values of $k$, 
and we found out that using the $k=5$ previous lines gives the best results. 

\spara{\lsi}.\ Bag-of-word models are not able to cope with synonymy nor polysemy. 
To enhance our model with such capabilities of use a simple latent semantic analysis (\lsi) approach.
As a preprocessing step, we remove stop words and words that appear less than 
three times.
Then we use our training data to form a line--term matrix and 
we compute a rank-100 approximation of this matrix. 
Each line is represented by a term vector and is projected on the low-rank 
matrix space.
The \lsi\ similarity between two lines is computed 
as the cosine similarity of their projected vectors. 

\spara{\nn}.\ 
Our last semantic feature is based on a neural language model. 
It is described in more detail in the next section.

\subsection{Methods}

In this section we present two building blocks of our method:
the neural language model used to incorporate semantic similarity with the \nn\ 
feature and the RankSVM method used to combine the various features.

\subsubsection{Neural language model} 
\label{sec:nn}

Our knowledge of all possible features can never be comprehensive, and
designing a solid extractor for features such as semantic or grammatical similarity would be 
extremely time-consuming. 
Thus, attempting to learn additional features appears a promising approach. 
Since neural networks can have a vast number of available parameters and 
can learn complex nonlinear mappings, we experiment with whether they could be 
used to automatically extract relevant features. 
To this end, we design a neural network that learns to use raw text sequences 
to predict the relevance of a candidate next line given previous lines.

In brief, our neural network starts by finding distributed (vector) 
representations for words. These are combined to distributed representations of 
lines, and further combined to vector representations of multiple lines, where 
the last line may be a real line, or a randomly sampled line. Based on this 
vector representation of text, our network learns to predict whether it believes 
that the last line is a suitable next line candidate or not.

\spara{Network architecture.} At the core of our predictor, we use 
multi-layered, fully-connected neural 
networks, trained with backpropagation. While the network structure was 
originally inspired by the work of Collobert et al.~\cite{collobert:2011b}, 
our model differs substantially from theirs. 
In particular, we have included new input transformations and our input 
consists of multiple sentences.

\begin{figure}[t]
	\includegraphics[width=\columnwidth]{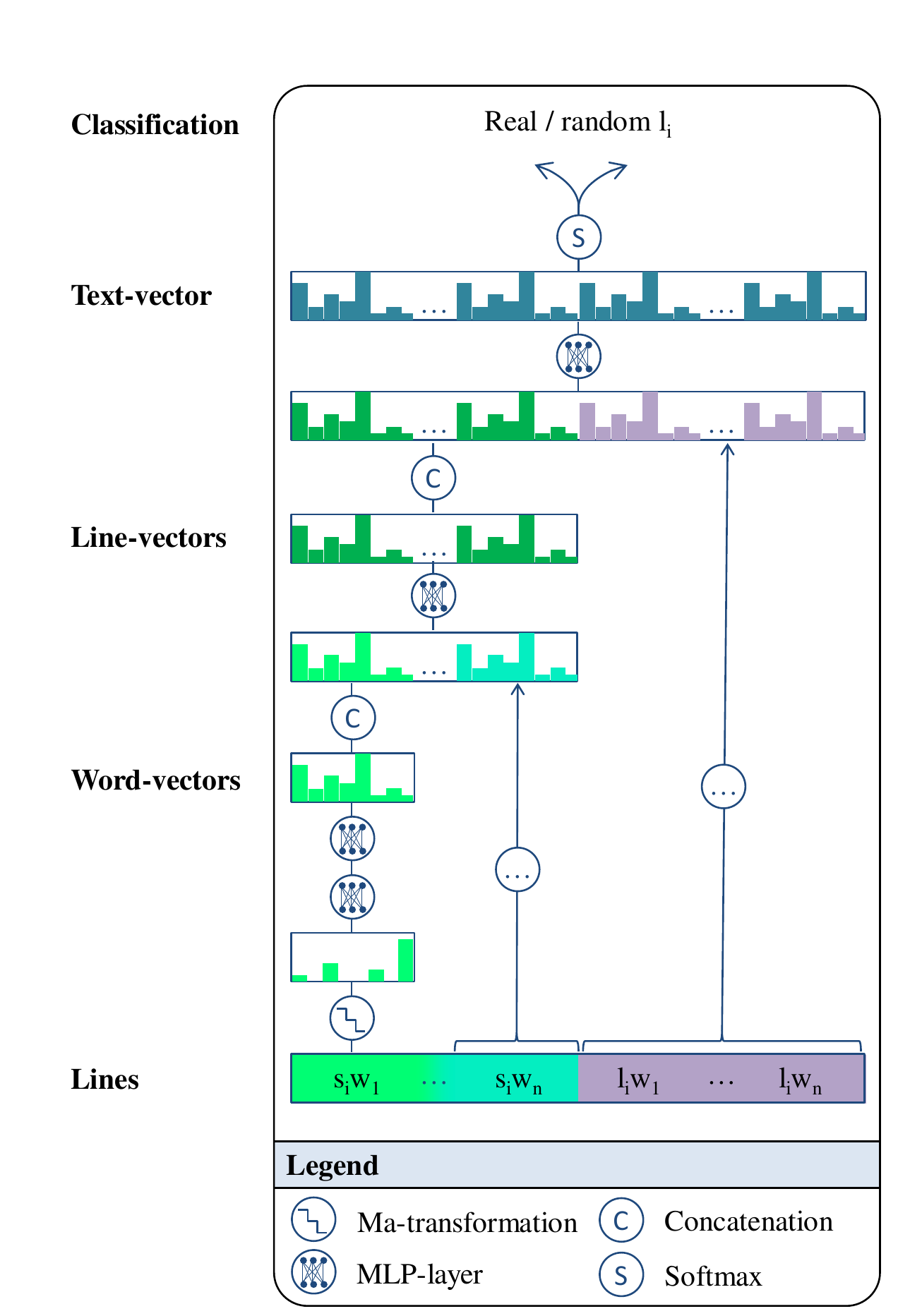}
	\caption{Network architecture}
	\label{fig:network_architecture}
\end{figure}

The structure of our model is illustrated in 
Figure~\ref{fig:network_architecture}.
We start by preprocessing all text to a format more easily handled by a 
neural network, removing most non-ascii characters and one-word lines, and 
stemming and lower-casing all words. Our choice of neural network architecture 
requires fixed line lengths, so we also remove words exceeding 13 words, and 
pad shorter lines. We build 
samples in the format: ``candidate next line (\(\ell_i\))''; ``previous lines 
(\(s_1, \ldots , s_m\)).'' For  lines with too few previous lines, we add 
paddings.

\begin{figure}[t]
	\centering
	\includegraphics[width=0.5\linewidth]{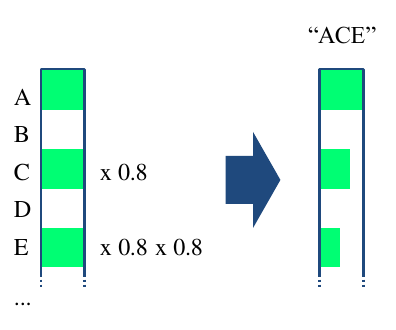}
	\caption{Simplified example of exponential moving-average transformation}
	\label{fig:moving_average}
\end{figure}

The first layers of the network are word-specific: they perform the same 
transformation for each word in a text, regardless of its position. For each word, we start with an exponential moving 
average transformation that turns a character sequence into a vector. A 
simplified example of this is shown in Figure~\ref{fig:moving_average} .\footnote{While we
achieve our best results using this transformation, a traditional `one-hot' vector approach
yields nearly as good results. A model that learns the transformation,
such as a recurrent neural network could also be tested in future work.} We use this transformation as it is, compared to one-hot encoding, more robust to different spellings, coping well for instance with slang and spelling errors.

The transformation works as follows: we create a zero-vector, with the length 
of possible characters. Starting with 
the first character in a word, we choose its corresponding location in the 
vector. We increment this value with a value 
proportional to the character's position in the word, counting from the 
beginning of the word. We get a word-representation
$\mathbf{w} = (w_a\, w_b \, \ldots \, w_z \, \ldots )^{T}$,
where
\[
w_a = \sum_{}^{} \frac{(1 - \alpha)^{c_a}}{Z},
\]
where \(c\) is the index of 
the character at hand, \(\alpha\) is a decay hyperparameter, and \(Z\) is a normalizer proportional to word length. For further details of this transformation, see \cite{Takala2016}.

To avoid always giving larger weight to the beginning of a word, we
also concatenate to the vector this transformation backwards, and a vector 
of character counts in the word.  
Following this transformation, we feed each word vector to a 
fully-connected neural network.

Our word-specific vectors are next concatenated in the word order to a single 
line-vector, and fed to a line-level network. Next, the 
line-vectors are concatenated, so that the candidate next-line is placed 
first, and preceding lines are then presented in their order of occurrence. 
This vector is fed to the final layer, and the output is fed to a 
softmax-function that outputs a binary prediction indicating whether it believes 
one line follows the next or not. In our ensemble model, we use the activation before the softmax, corresponding to the confidence the network has in a line being the next
in the lyrics.

\spara{Training.} For training the neural network, we retrieve a 
list of previous lines for each 
of the lyrics lines. These lyrics are fed 
to the neural model in small batches, and a corresponding gradient descent 
update is performed on the weights of the network. To give our network
negative examples, we follow an approach similar to \cite{collobert:2011b}, 
and generate fake line examples by choosing for every 
second example the candidate line uniformly at random from all lyrics.

A set of technical choices is necessary for constructing and training our 
network. We use two word-specific neural layers (500 neurons each), one 
line-specific layer (256 neurons, and one final layer (256 neurons). All 
layers have rectified linear units as the activation function. Our minibatch 
size is 10, we use the adaptive learning rate Adadelta \cite{adadelta}, and we 
regularize the network with 10\% dropout \cite{JMLR:v15:srivastava14a}. We 
train the network for 20 epochs on a GPU machine, taking advantage of the 
Theano library \cite{bergstra2010theano}. Analyzing hyperparameters, we find that we can improve results especially by going from one previous line to a moderately large context (5 previous lines), and by using a larger input window, i.e. giving the network more words from each line.

\spara{Manual evaluation.} To get some understanding of what our neural network has 
learnt, we manually
analyze 25 random sentences where a random candidate line was falsely classified as
being the real next line. For 13
of the lines, we find it difficult to distinguish whether the real or the random line
should follow. This often relates to the rapper changing the topic, especially
when moving from one verse to another. For the remaining 12 lines, we notice 
that the neural network has not succeeded in identifying five instances with a 
rhyme, three instances with repeating words, two instances with consistent 
sentence styles (lines having always e.g., one sentence per
line), one instance with consistent grammar (a sentence continues to next 
line), and one instance where the pronunciation of words is very similar the 
previous and the next line.
In order to detect the rhymes, the network would need to learn a phonetic 
representation of the words, but for this problem we have developed other 
features presented in Section~\ref{sec:feats}.

\subsubsection{Ranking candidate lines} \label{sec:ranksvm}


We take existing rap lyrics as the ground truth for the next line prediction 
problem. The lyrics are transformed into preferences
\begin{equation*}
 s_{m+1} \succ_B \ell_i,
\end{equation*}
which are read as ``$s_{m+1}$ (the true next line) is preferred over $\ell_i$ (a 
randomly sampled line from the set of all available lines) in the context of 
$B$ (the preceding lines)''. When sampling $\ell_i$ we ensure that $\ell_i \neq 
s_{m+1}$.
Then we extract features $\match{B, \ell}$ listed in Section~\ref{sec:feats} 
to describe the match between the preceding lines and a candidate next line.

The RankSVM algorithm \cite{joachims2002} takes this type of data as input 
and tries to learn a linear model which gives relevance scores for the 
candidate lines. These relevance scores should respect the preference relations 
given as the training data for the algorithm. The relevance scores are given by
\begin{equation} \label{eq:ranksvm}
 r(B, \ell) = \w^T \match{B,\ell},
\end{equation}
where $\w$ is a weight vector of the same length as the number of features.

The advantage of having a simple linear model is that it can be trained very 
efficiently. We employ the $\text{SVM}^{\text{rank}}$ software for the 
learning~\cite{joachims2006}. Furthermore, we can interpret the weights as importance values for 
the features. Therefore, when the model is later applied to lyrics generation, 
the user could manually adjust the weights, if he or she, for instance, prefers 
to have lyrics where the end rhyme plays a larger/smaller role.

\subsection{Empirical next line prediction evaluation} \label{sec:feat_res}

Our experimental setup is the following. We split the lyrics by artist randomly into 
training (50\%), validation (25\%), and test (25\%). The RankSVM model 
and the neural network model are learned using the training set, while varying 
a trade-off parameter $C$  
among $\{1$,\ $10^1$, $10^2$, $10^3$, $10^4$, $10^5\}$ to find the value 
which 
maximizes the performance on the validation set. The final performance is 
evaluated on the unseen test set.

We use a candidate set containing the true next line and 299 randomly chosen 
lines. The performance is measured using mean rank, mean reciprocal rank 
(reciprocal value of the harmonic mean of ranks), and recall at $N$ where the 
value of $N$ is varied.
The results are computed based on a random sample of 10\,000 queries.

Table~\ref{tab:res} shows the test performance for different feature sets. Each 
feature alone yields a mean rank of $<150$ and thus carries some predictive 
power. 
The best individual feature with respect to the mean rank is the output of the neural 
network model. However, if we look at recall at 1 (the probability to rank the 
true next line first), the best performance, 7.7\%, is obtained by \erhyme 
which works very well in some cases but is in overall inferior to \nn since in 
some cases consecutive lines do not rhyme at all.
Combining all features, we achieve a mean rank of 60.8 and can pick the true 
next line with 16.9\% accuracy. The probability to pick the true next line at 
random is only 0.3\%. 
The features are combined by taking a linear combination of their values 
according to Equation~(\ref{eq:ranksvm}).

To enable the generation of rap lyrics in real time, we also tested 
employing only the features which are fast to evaluate, i.e., \fastfeats = 
\linelength + \erhyme + \eerhyme + \bow + \bowfive, and \fastfeatsnn = 
\fastfeats + \nn. With the latter feature set, the performance is almost 
identical to using the full feature set and even \fastfeats works relatively 
well.

%

\begin{table}[t]
\caption{Next line prediction results for $k=300$ candidate 
lines. MRR stands for mean reciprocal rank and Rec@N for recall 
when retrieving top $N$ lines.}\label{tab:res}
\begin{center}
\resizebox{\columnwidth}{!}{
\addtolength{\tabcolsep}{-3pt}
\begin{tabular}{lccccccc}
\toprule
Feature(s) & Mean rank & MRR & Rec@1 & Rec@5 & Rec@30 & Rec@150\\
\midrule

Random & 150.5 & 0.021 & 0.003 & 0.017 & 0.010 & 0.500\\
\linelength & 117.6 & 0.030 & 0.002 & 0.029 & 0.177 & 0.657\\
\erhyme & 103.2 & \textbf{0.140} & \textbf{0.077} & \textbf{0.181} & 
\textbf{0.344} & 0.480\\
\eerhyme & 126.0 & 0.075 & 0.037 & 0.092 & 0.205 & 0.347\\
\intrhyme & 123.3 & 0.047 & 0.016 & 0.055 & 0.190 & 0.604\\
\bow & 112.4 & 0.116 & 0.074 & 0.138 & 0.280 & 0.516\\
\bowfive & 99.1 & 0.110 & 0.065 & 0.129 & 0.314 & 0.708\\
\lsi & 111.3 & 0.089 & 0.051 & 0.107 & 0.262 & 0.662\\
\nn & \textbf{84.7} & 0.067 & 0.020 & 0.083 & 0.319 & \textbf{0.793}\\
\midrule
FastFeats & 73.5 & 0.224 & 0.160 & 0.272 & 0.476 & 0.802\\
FastFeatsNN5 & 61.2 & \textbf{0.244} & \textbf{0.172} & \textbf{0.306} & 
0.524 & 0.853\\
\midrule
\textbf{All features} & \textbf{60.8} & 0.243 & 0.169 & 0.304 & \textbf{0.527} 
& \textbf{0.855}\\
\bottomrule
\end{tabular}
\addtolength{\tabcolsep}{3pt}
}
\end{center}
\end{table}


\newpage
\section{Lyrics generation}

\subsection{Overview}

The lyrics generation is based on the idea of selecting the most relevant line 
given the previous lines, which is repeated until the whole lyrics have 
been generated. Our algorithm \deepbeat is summarized in Algorithm~\ref{alg:gen}.

\begin{algorithm}[t]
\caption{Lyrics generation algorithm \deepbeat.}
\label{alg:gen}

\KwIn{Seed line $\ell_1$, length of the lyrics $n$.}
\KwOut{Lyrics $L = \left(\ell_1, \ell_2, \ldots, \ell_n\right)$.}
$L[1] \leftarrow \ell_1$ \tcp*[r]{\scriptsize{Initialize a list of lines.}}
\For{$i\leftarrow 2$ \KwTo $n$}{
  $C \leftarrow \mathtt{retrieve\_candidates}(L[i-1])$ 
\tcp*[r]{\scriptsize{Sec.~\ref{sec:retrieval}}}
  $\hat{c} \leftarrow NaN$\;
  \ForEach{$c \in C$}{
    \tcc{\scriptsize{Check relevance and feasibility of the candidate.}}
    \If{$\rel{c, L} > \rel{\hat{c}, L}\, \& \, \mathtt{rhyme\_ok}(c, L)$}{
      $\hat{c} \leftarrow c$\;
    }
  }
  $L[i] \leftarrow \hat{c}$\;
}
\Return $L$\;
\end{algorithm}

The relevance scores for candidate lines are computed using the RankSVM 
algorithm described in Section~\ref{sec:nextline}. Instead of selecting the 
candidate with strictly the highest relevance score, we filter some candidates 
according to \mbox{\texttt{rhyme\_ok()}} function. It checks that consecutive 
lines are 
not from the same song and that they do not end with the same words. Although 
rhyming the same words can be observed in some contemporary rap lyrics, it has 
not always been considered a valid technique \cite{edwards2009}. While it is 
straightforward to produce long ``rhymes'' by repeating the same phrases, it can 
lead to uninteresting lyrics.

\subsection{Online demo} \label{sec:demo}

An online demo for the lyrics generation algorithm is available at \website. 
This web tool was built in order to (1)~make the generator available to the 
public, (2)~provide the users easy ways of customizing the generated
lyrics, and (3)~collect limited usage logs in order to evaluate and improve the 
algorithm. The website was launched in November 2015 and as of June 2016 it 
has been visited by more than 42\,000 users.

After the initial launch of the project, we started collaborating with some 
musicians to record the first songs written by \deepbeat,\footnote{The first 
music video 
is available at: \\ \url{https://youtu.be/Js0HYmH31ko} \\
More about the collaboration can be read at: \\
\url{https://howwegettonext.com/deepbeat-what-happens-}
\url{when-a-robot-writes-rhymes-for-rappers-77d07c406ff5}} which showed us the 
importance of giving the user sufficient customization capabilities instead of 
merely outputting complete lyrics. With this in mind, we designed the online 
demo so that users can: (1)~define 
keywords  that must appear in the generated lyrics; (2)~ask the algorithm to 
give suggestions for the next line and pick the best suggestion manually; and 
(3)~write some of the 
lines by themselves. A user can, for example, write the first line by herself 
and let the algorithm generate the remaining lines. An interesting mode of 
usage we noticed some users adopting is to write every other line by yourself 
and generate every other.

When the user generates lyrics line-by-line, asking for suggestions from \deepbeat, 
we log the selected lines. In Section~\ref{sec:experiment}, we show how to 
evaluate the algorithm using the log data. Conveniently, we can also use the 
logs to refine the learned models employing the RankSVM approach as done in 
\cite{joachims2002}.

\subsubsection{Performance optimization} \label{sec:retrieval}

\spara{Candidate set retrieval.} Results in Table~\ref{tab:res} show that 
\erhyme alone is a good predictor. Therefore, we define the set of candidate 
next lines as the 300 best rhyming lines instead of 300 random lines. This 
candidate set has to be retrieved quickly without evaluating each of the $n_l = 
583\,669$ lines in our database.
Conveniently, this can be formulated as the problem of 
finding $k$ strings with the longest common prefix with respect to a query 
string, as follows
\squishlist
 \item[1.] Compute a phonetic transcription for each line.
 \item[2.] Remove all but vowel phonemes.
 \item[3.] Reverse phoneme strings.
\squishend

We solve the longest common prefix problem by first, sorting the phoneme string 
as a preprocessing step, second, employing binary search\footnote{The candidate 
set could be found even faster by building a trie data structure of 
the phoneme strings. However, in our experiments, the binary search approach 
was fast enough and more memory-efficient.} to find the 
line with the longest common prefix ($\ell_\textmd{long}$), and third, taking the 
$k-1$ lines with the longest common prefix around $\ell_\textmd{long}$. 
The computational complexity of this approach is $\mathcal{O}(\log n_l + k)$. 
For the online demo, we have set $k=300$.

\spara{Feature selection.} By default, \website employs the \fastfeats feature 
set by which the generation of an 8-line verse takes about 0.3 seconds. The user 
can additionally enable the \nn feature in which case the algorithm will 
retrieve the top-30 lines based on \fastfeats and then rerank the top lines 
based on \fastfeatsnn. We only evaluate 30 lines using \nn since \nn is much 
heavier to compute than the other features (it increases the generation 
time to about 35 seconds per 8-line verse) and since recall at 30 is only 3.8 
percentage points lower for \fastfeats compared to \fastfeatsnn. The \nn 
feature could be used more heavily by acquiring a server which enables GPU 
computation or via parallelization (different candidate lines can be evaluated 
independently).


\subsection{Empirical evaluation of generated lyrics} \label{sec:res}

The lyrics generated by \deepbeat are evaluated by the rhyme density measure, 
introduced in Section~\ref{sec:rhymedensity}, and by measuring the correlation 
between relevance scores assigned by \deepbeat and human preferences recorded 
via \website.

\subsubsection{Rhyme density of DeepBeat}

We ran a single job to generate a hundred 16-bar verses with random seed lines.
%
%
%
%
A randomly selected example verse from this set is shown in 
Appendix~\ref{sec:verses}.
%
The rhyme density for the hundred verses is 1.431, which is, quite 
remarkably, 21\% higher than the rhyme density of the top ranked human rapper, 
Inspectah Deck.

One particularly long multisyllabic 
rhyme created by the algorithm is given below (the rhyming part is highlighted)
\setlength{\tabcolsep}{2pt}
\begin{center}
\begin{tabular}{rcl}
\small{``Drink and drown} & \small{{\bf in my }} & \small{{\bf 
own iniquity}}\\
\small{Never smile style} & \small{{\bf is wild}} & 
\small{{\bf only grin strictly}}''



\end{tabular}
\end{center}
\setlength{\tabcolsep}{6pt} 
In this example, the first line is from the song \emph{Rap Game} by \emph{D-12} 
and the latter from \emph{I Don't Give a F**k} by \emph{AZ}. The rhyme consists 
of nine consecutive matching vowel phonemes.

\subsubsection{Online experiment} \label{sec:experiment}

In order to evaluate the algorithm, we performed an online experiment using the 
demo. The idea was to employ an approach which is used for 
optimizing search engines \cite{joachims2002} where the clicked search result 
$r_i$ is logged and the following pairwise preferences are extracted: $r_i 
\succ r_j,\, j=1,\ldots,i-1$ (the lines below the selected line are ignored 
since we cannot assume that the user has evaluated those). The objective is to 
learn a ranking model which assigns relevance scores that agree with the 
extracted preferences.

At \website, when a user clicks ``Suggest Rhyming Line'' button, 20 suggested 
candidate next lines are shown to the user. We wanted to see how often the line 
selected by the user was assigned a higher score than the lines above the 
selection.
Our hypothesis was that the larger the absolute difference between the 
algorithm-assigned relevance scores of two lines was, the more likely the user 
would pick the line preferred by the algorithm.

In the initial data we collected through the website, we noticed that users 
are more likely to select a line the higher it appears on the list of 
suggestions. Furthermore, the users tend to prefer the fifth line since it often 
appears in the same location of the screen as the ``Suggest Rhyming Line'' 
button, so if a user is just playing around with the tool without putting much 
thought to the content of the suggestions, the user is likely to select the 
fifth suggestion. It is very challenging to get rid of this type of biases 
entirely when conducting an uncontrolled experiment in the wild but to mitigate 
the biases, we shuffled the order of the suggested lines and removed the first 
three selections of each user since we assumed that in the beginning users 
are more likely to play with the tool without thinking too much. Moreover, we 
wanted to create more variability among the suggestions, as the top 30 might 
be almost equally good, so we defined the set of suggested lines as the lines 
ranked: 1--14., 298--300., and three randomly picked lines from range 15--297.

To avoid degrading the usability of the tool too much, we only applied the 
aforementioned manipulations when \nn was not enabled by the user. However, 
we also stored the text of the selected line and the previous lines to enable 
the computation of the relevance scores with \fastfeatsnn as a post-processing 
step. In total, this experiment resulted in 34\,757 pairwise preferences from 
1\,549 users.\footnote{An anonymized version of the dataset including the 
scores assigned by \deepbeat is available at: \\ \url{https://github.com/ekQ/dopelearning}}

\begin{figure}[t]
  \centering
  \includegraphics[width=\columnwidth]{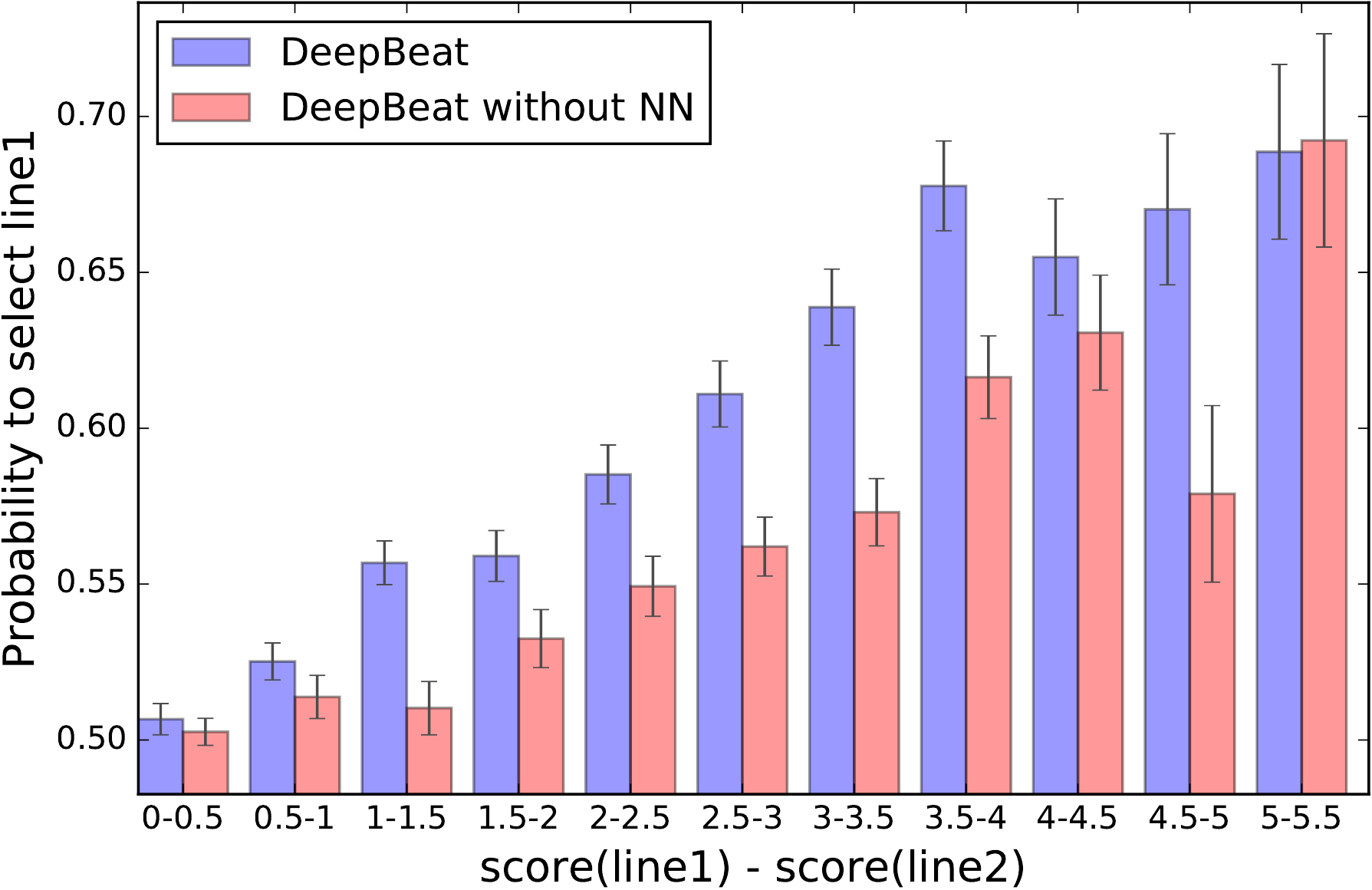}
  \caption{Probability of a \website user to select a line with a higher score from a pair of lines
  given the (binned) score difference of the lines. User preferences correlate with
  the scores assigned by DeepBeat.}
  \label{fig:web}
\end{figure}

The results of the experiment are shown in Figure~\ref{fig:web}. They 
confirm our hypothesis that the higher the difference between the relevance 
scores of two lines the more likely the users select the line which is 
evaluated more suitable by the algorithm. Furthermore, including the \nn 
feature 
improves the evaluations, which can be seen by studying the difference of 
the two data series in the figure. We may conclude that the learned RankSVM 
model generalizes and is 
able to successfully learn human preferences from existing rap lyrics, and that 
the developed deep neural network is an important component of the model.

\section{Discussion}

According to Boden \cite{Boden2004}, creativity is the ability to come up with 
ideas or artifacts that are (1)~new, (2)~surprising, and (3)~valuable.
Here, produced lyrics as a whole are novel by construction as
the lines of the lyrics are picked from different original lyrics,
even if individual lines are not novel.
Lyrics produced with our method are likely to be at least as
surprising as the original lyrics.

The (aesthetic) value of rap lyrics is
more difficult to estimate objectively.
Obvious factors contributing to the value are
poetic properties of the lyrics, especially its rhythm and rhyme,
quality of the language, and the meaning or message of the lyrics.
Rhyme and rhythm can be controlled with relative ease,
even outperforming human rappers, as we have demonstrated in our experiments.
The quality of individual lines is exactly as good as the dataset used.

The meaning or semantics is the hardest part for computational generation.
We have applied standard bag-of-words and LSA methods and additionally 
introduced a deep neural network model in order to capture the semantics of the 
lyrics. The importance of these features has been proven by the experimental 
results for the next line prediction problem and an online experiment. The
features together contribute towards the semantic coherence of the 
produced lyrics even if full control over the meaning is still missing.

The task of predicting the next line can be challenging even for a human, and 
our model performs relatively well on this task. We have used our models for 
lyrics prediction, but we see that components that understand semantics could be 
very relevant also to other text processing tasks, for instance conversation 
prediction.

This work opens several lines for future work. It would be interesting to study 
automatic creation of story lines by analyzing existing rap songs and of novel 
lines by modifying existing lines or creating them from scratch. Even more, it 
would be exciting to have a 
fully automatic rap bot which would generate lyrics 
and rap them synthetically based on some input it receives from the outside 
world. Alternatively, the findings of this paper could be transferred to other 
text processing tasks, such as conversation prediction which could carry 
significant business potential when applied to tasks like customer service 
automation.

\section{Conclusions}

We developed \deepbeat, an algorithm for rap lyrics generation. Lyrics 
generation was formulated as an information retrieval task where the objective 
is to find the most relevant next line given the previous lines which are 
considered as the query. The algorithm extracts three types of features of the 
lyrics---rhyme, structural, and semantic features---and combines them 
by employing the \ranksvm algorithm. For the semantic features, we developed a 
deep neural network model, which was the single best predictor for the 
relevance of a line.

We quantitatively evaluated the algorithm with three measures. First, we 
evaluated prediction performance by measuring how well the algorithm
predicts the next line of an existing rap song. 
The true next line was identified among 299 randomly selected lines with an 
accuracy of 17\%, i.e., over 50 times more likely than by random, and it was 
ranked in the top 30 with 53\% accuracy.
Second, we introduced a rhyme density measure and showed that \deepbeat 
outperforms the top human rappers by 21\% in terms of length and frequency of 
the rhymes in the produced lyrics. The validity of the rhyme density measure was 
assessed by conducting a human experiment which showed that the measure 
correlates with a rapper's own notion of technically skilled lyrics.
Third, the rap lyrics generator was deployed as a web tool (\website) and the 
analysis of its usage logs showed that machine evaluations of candidate next 
lines correlate with user preferences.

\section*{Acknowledgments}

We would like to thank Stephen Fenech for developing the front end for 
\website and Jelena Luketina, Miquel Perell\'{o} Nieto, and Vikram 
Kamath for useful comments on the manuscript.

%
\bibliographystyle{abbrv}
\bibliography{dopelearning}  

\appendix
\section{Sample Verses} \label{sec:verses}

Table~\ref{tab:sample2} shows an example of a generated verse. It was randomly 
selected from a set of a hundred verses, excluding the verses with profane 
language. The other verses are available at: 
\url{https://github.com/ekQ/dopelearning}

\input{sample2.tex}

\balancecolumns

%

%
\end{document}

%% file: defines.tex
\usepackage{suffix}

\usepackage{color}

\renewcommand\sf[1]{\textsf{#1}}

\newcommand{\nextline}{{\sf NextLine}\xspace}
\newcommand{\erhyme}{\sf{EndRhyme}\xspace}
\newcommand{\eerhyme}{\sf{EndRhyme-1}\xspace}
\newcommand{\intrhyme}{\sf{OtherRhyme}\xspace}
\newcommand{\linelength}{\sf{LineLength}\xspace}
\newcommand{\bow}{\sf{BOW}\xspace}
\newcommand{\bowfive}{\sf{BOW5}\xspace}
\newcommand{\nn}{\sf{NN5}\xspace}
\newcommand{\lsi}{\sf{LSA}\xspace}
\newcommand{\fastfeats}{\sf{FastFeats}\xspace}
\newcommand{\fastfeatsnn}{\sf{FastFeatsNN5}\xspace}
\newcommand{\ranksvm}{\sf{RankSVM}\xspace}
\newcommand{\deepbeat}{\sf{DeepBeat}\xspace}
\newcommand{\website}{\mbox{\texttt{deepbeat.org}}\xspace}

\newcommand{\len}[1]{{\textmd{len}\left(#1\right)}}
\newcommand{\rel}[1]{{\textmd{rel}\left(#1\right)}}

\newcommand{\w}{{\ensuremath{\mathbf{w}}}}
\newcommand{\match}[1]{{\ensuremath{\mathbf{\Phi}\left(#1\right)}}}

\newcommand{\spara}[1]{\smallskip\noindent{\bf{#1}}}

\newcommand{\fpr}[1]{\mathopen{}\left(#1\right)}

\newcommand{\abs}[1]{{\left|#1\right|}}

\DeclareRobustCommand{\dispfunc}[2]{%
  \ensuremath{%
  \ifthenelse{\equal{#2}{}}%
    {{#1}}%
    {{#1}\fpr{#2}}}}

\newcommand{\dd}{\ensuremath{K}}

\newcommand{\dist}[1][(p, q)]{\def\ArgI{{#1}}\distRelay}
\newcommand{\distRelay}[2][]{\dispfunc{\dd^{\ArgI}_{#1}}{#2}}
\WithSuffix\newcommand\dist*[2][]{\dist[][#1]{#2}}

\newtheorem{problem}{Problem}

\newcommand{\squishlist}{\begin{list}{$\bullet$}
  { \setlength{\itemsep}{0pt}
     \setlength{\parsep}{3pt}
     \setlength{\topsep}{3pt}
     \setlength{\partopsep}{0pt}
     \setlength{\leftmargin}{1.5em}
     \setlength{\labelwidth}{1em}
     \setlength{\labelsep}{0.5em} } }
\newcommand{\squishend}{
\end{list}  }

\SetKwComment{tcpas}{\{}{\}}
\SetCommentSty{textnormal}
\SetArgSty{textnormal}
\SetKw{False}{false}
\SetKw{True}{true}
\SetKw{Null}{null}
\SetKwInOut{Output}{output}
\SetKwInOut{Input}{input}
\SetKw{AND}{and}
\SetKw{OR}{or}
\SetKw{Break}{break}

\pgfdeclarelayer{background}
\pgfdeclarelayer{foreground}
\pgfsetlayers{background,main,foreground}

\definecolor{yafaxiscolor}{rgb}{0.3, 0.3, 0.3}

\definecolor{yafcolor1}{rgb}{0.4, 0.165, 0.553}
\definecolor{yafcolor2}{rgb}{0.949, 0.482, 0.216}
\definecolor{yafcolor3}{rgb}{0.47, 0.549, 0.306}
\definecolor{yafcolor4}{rgb}{0.925, 0.165, 0.224}
\definecolor{yafcolor5}{rgb}{0.141, 0.345, 0.643}
\definecolor{yafcolor6}{rgb}{0.965, 0.933, 0.267}
\definecolor{yafcolor7}{rgb}{0.627, 0.118, 0.165}
\definecolor{yafcolor8}{rgb}{0.878, 0.475, 0.686}

\tikzstyle{exnode} = [inner sep = 1pt]
\tikzstyle{exedge} = [yafcolor5, draw, thick, >=latex, ->]
\tikzstyle{toynode} = [fill = yafcolor5, circle, inner sep = 1.5pt]
\tikzstyle{toyedge} = [draw = black, >=latex, ->, opacity = 0.5]

%% file: sample2.tex
\begin{table}[!htb]
\caption{A randomly selected verse from the set of 100 generated lyrics.} 
\label{tab:sample2}
\begin{center}
\resizebox{\columnwidth}{!}{
\begin{tabular}{ll}
\small{Everybody got one} & {\color{darkgray} \tiny{(2 Chainz - Extremely 
Blessed)}} \\
\small{And all the pretty mommies want some} & {\color{darkgray} \tiny{(Mos Def 
- Undeniable)}} \\
\small{And what i told you all was} & {\color{darkgray} \tiny{(Lil Wayne - 
Welcome Back)}} \\
\small{But you need to stay such do not touch} & {\color{darkgray} 
\tiny{(Common 
- Heidi Hoe)}} \vspace{0.1cm} \\
\small{They really do not want you to vote} & {\color{darkgray} \tiny{(KRS One 
- 
The Mind)}} \\
\small{what do you condone} & {\color{darkgray} \tiny{(Cam'ron - Bubble 
Music)}} \\
\small{Music make you lose control} & {\color{darkgray} \tiny{(Missy Elliot - 
Lose Control)}} \\
\small{What you need is right here ahh oh} & {\color{darkgray} \tiny{(Wiz 
Khalifa - Right Here)}} \vspace{0.1cm} \\
\small{This is for you and me} & {\color{darkgray} \tiny{(Missy Elliot - 
Hit Em Wit Da Hee)}} \\
\small{I had to dedicate this song to you Mami} & {\color{darkgray} \tiny{(Fat 
Joe - Bendicion Mami)}} \\
\small{Now I see how you can be} & {\color{darkgray} \tiny{(Lil Wayne - How To 
Hate)}} \\
\small{I see u smiling i kno u hattig} & {\color{darkgray} \tiny{(Wiz Khalifa - 
Damn Thing)}} \vspace{0.1cm} \\
\small{Best I Eva Had x4} & {\color{darkgray} \tiny{(Nicki Minaj - Best I Ever 
Had)}} \\
\small{That I had to pay for} & {\color{darkgray} \tiny{(Ice Cube - X 
Bitches)}} 
\\
\small{Do I have the right to take yours} & {\color{darkgray} \tiny{(Common - 
Retrospect For Life)}} \\
\small{Trying to stay warm} & {\color{darkgray} \tiny{(Everlast - 2 Pieces Of 
Drama)}} \\
\end{tabular}
}
\end{center}
\end{table}